\documentclass[preprint,12pt]{elsarticle}

\pdfoutput=1
\usepackage{amssymb}
\usepackage{float}
\usepackage{epstopdf}
\usepackage{algorithm,algcompatible,amsmath}
\usepackage{xcolor}
\newcommand\mynotes[1]{\textcolor{red}{#1}}
\newcommand\tgknotes[1]{\textcolor{green}{#1}}
\renewcommand\mynotes[1]{} 
\renewcommand\tgknotes[1]{} 

\journal{Neurocomputing: Special Issue on MORL: Theory and applications: }

\begin{document}

\begin{frontmatter}



\title{Identification and Off-Policy Learning of Multiple Objectives Using Adaptive Clustering}


\author{Thommen George Karimpanal, Erik Wilhelm}
\ead{thommen\_george@mymail.sutd.edu.sg, erikwilhelm@sutd.edu.sg}
\address{Singapore University of Technology and Design, 8 Somapah Road, Singapore 487372}

\begin{abstract}
In this work, we present a methodology that enables an agent to make efficient use of its exploratory actions by autonomously identifying possible objectives in its environment and learning them in parallel. The identification of objectives is achieved using an online and unsupervised adaptive clustering algorithm. The identified objectives are learned (at least partially) in parallel using $Q-$learning. Using a simulated agent and environment, it is shown that the converged or partially converged value function weights resulting from off-policy learning can be used to accumulate knowledge about multiple objectives without any additional exploration. We claim that the proposed approach could be useful in scenarios where the objectives are initially unknown or in real world scenarios where exploration is typically a time and energy intensive process. The implications and possible extensions of this work are also briefly discussed.

\end{abstract}


\end{frontmatter}



\section{Introduction}
\label{intro}

Intelligent agents are characterized by their abilities to learn from and adapt to their environments with the objective of performing specific tasks. 
  Very often, in reinforcement learning \cite{sutton1998reinforcement}, and in machine learning in general, algorithms are structured to be able to fulfill one specific objective, usually specified in terms of a particular region in the feature space that is associated with a high reward.  In general, environments are likely to contain multiple features, and different regions in the feature space may specify different objectives that could be assigned to the agent to learn. In real-world scenarios, however, the ability to efficiently learn more than one objective during a single deployment could drastically improve the agent’'s usefulness. In order to achieve this, the agent would need to be aware of regions in the feature space that could possibly play a role in its future tasks.

Embodied artificial agents or intelligent robots are typically equipped with a variety of sensors that enable it to detect characteristic features in its environment. In the context of reinforcement learning, when such an agent is placed in an unknown environment and is assigned an objective, it carries out some form of exploratory behavior in order to first discover a region in the feature space that fulfills this objective. Further exploratory actions may help improve its value function estimates, which in turn lead to improved policies to achieve the objective. We shall refer to this original task as the \emph{primary objective}, and to its associated feature vector as the \emph{primary objective feature vector} ($\vec{\psi}$).  During exploration, it is likely that the agent comes across other `interesting' regions which contain features that stand out with respect to the agent's history of experiences. We shall refer to these regions of the feature space as \emph{secondary objectives}, and to the associated feature vectors as  \emph{secondary objective feature vectors} ($\vec{\phi}$). Although these regions could be of interest to the agent for future tasks (which are currently unknown), they may be irrelevant to the task at hand. Hence, it is justified for the agent to ignore them and continue performing value function updates for the primary objective assigned to it. 
 
 However, the agent's future tasks may not remain the same and a new task assigned to it may correspond to a particular combination of features that it encountered while learning policies for the primary objective. In such a case, the fact that this region in the feature space had been previously encountered cannot be leveraged since they were not relevant to the agent at that point of time, and were hence ignored.
 

The above mentioned approach would result in a considerable amount of wasteful exploration. This is because each new task assigned to the agent would require a fresh phase of discovery and learning of the associated feature vector and value functions respectively.  
A more efficient approach would be to keep track of possible secondary objectives and learn them in parallel using off-policy methods \cite{precup2001off,sutton1998reinforcement}. In the context of off-policy learning, this can be done by treating the policies corresponding to the secondary objectives as target policies, and learning them while executing the behavior policy which is dictated by the primary objective. Depending on the objectives, the actions executed by the behavior policy may not be optimal with respect to the secondary objectives. However, using off-policy learning, it is possible to at least partially learn the value functions for the secondary objectives, thereby significantly improving the efficiency of exploration. In applications such as robotics where exploration is known to be costly in terms of time, energy and other factors, such an approach could prove to be practical.
 
 In this work, we present a framework in which an unsupervised, adaptive clustering algorithm is designed and used to cluster regions of the feature space into different groups based on the similarity of their associated features. Off-policy methods are used to simultaneously learn target policies corresponding to these clusters, each of which is treated as a secondary objective. The clustering of features occurs as and when they are seen by the agent while learning the primary task. The value function updates can be performed using suitable off-policy methods, namely, tabular $Q-$ learning, $Q-$ $\lambda$ \cite{watkins1989learning} or other more recent off-policy methods \cite{geist2014off} such as off-policy LSTD($\lambda$) \cite{yu2010convergence,lagoudakis2003least}, off-policy TD($\lambda$) \cite{precup2000eligibility,precup2001off}, GQ($\lambda$) \cite{maei2010gq} etc., 
The results presented here, however, correspond to the $Q-\lambda$ algorithm.  

The primary objectives have an influence on the discovery and learning of the secondary objectives, but only through its behavior policy. As long as the agent executes some exploratory actions while learning to perform its primary task, secondary objectives can be discovered and at least partially be learned. In fact, even a purely exploratory policy can be used. These aspects are discussed in further detail in Section \ref{results}. 
 
 Ideally, our approach would obviate the need for a fresh phase of discovery and learning when the objective is changed. However, the aim  here is not to learn all the secondary objectives perfectly, but to identify them via the adaptive clustering algorithm, and learn them at least partially through off-policy learning. Doing so could provide the agent with a good initialization of value function weights so that optimal policies for the identified possible objectives could be learned in the future, if needed.
 

\section{Background}
\label{background}
Reinforcement learning deals with developing strategies for an agent to act in its environment with the objective of maximizing the expected value of a scalar reward. Most research in reinforcement learning is based on the formalism of Markov Decision Processes (MDPs) \cite{Puterman:1994:MDP:528623}.
In this framework, an agent in state $s\in \mathcal S$  takes an action $a\in \mathcal A$ to transition into a new state $s'$ with a probability $P(s,a,s')$. At each state, the agent receives a scalar reward $R(s,a)$. All reinforcement learning methods can be thought of as ways to maximize the expected reward accumulated over time as the agent interacts with the environment. The outcome of these methods is a mapping from states to actions, referred to as a policy. If the learning agent learns the value function for the policy being executed, it is referred to as \emph{on-policy} learning, and if it learns the value function for an objective irrespective of the policy being executed, it is called \emph{off-policy} learning.

In this work, our goal is to identify secondary objectives and learn their corresponding policies in parallel while the agent executes its behavior policy based on its primary objective.  Hence,  \emph{off-policy} learning methods are a natural choice for the stated goal. We use the $Q-\lambda$ algorithm, which is an extension of tabular $Q-$ learning that is suitable for application in continuous state spaces.  The update equation for the tabular case is shown in equation \ref{Qlearning }. 


\begin{equation}\label{Qlearning }
Q(s,a)\leftarrow Q(s,a)+\alpha[R(s,a)+\gamma max_{a'}Q(s',a')-Q(s,a)]
\end{equation}
where $Q(s,a)$ is the $Q-$value corresponding to state $s$ and action $a$. $s'$ is the next state, and $a'$ is a bound variable that can represent any action in the action space $\mathcal A$. $\alpha$ is the learning rate and $\gamma$ is the discount factor.

The $Q-\lambda$ algorithm performs a similar, but more involved update with weight vectors, and involves the use of eligibility traces \cite{sutton1988learning}. Here, replacing traces are used for the $Q-\lambda$ updates \cite{singh1996reinforcement}. 
The update equations for the $Q-\lambda$ algorithm are mentioned below:
\begin{equation}\label{TD_update}
\delta\leftarrow \delta+\gamma max_{a'}Q(s',a')
\end{equation}
\begin{equation}\label{Qlambda-weight_update}
w\leftarrow w+\alpha\delta e
\end{equation}
\begin{equation}\label{trace_update}
e\leftarrow \gamma\lambda e
\end{equation}
where $w$ is the weight vector, $e$ is the eligibility trace vector, $\lambda$ is the trace decay rate parameter and $\delta$ is defined as: 
\begin{equation}\label{Qlambda-TDerror}
\delta=R(s,a)-Q(s,a)
\end{equation}
The elements of the eligibility trace vector (replacing traces) are initialized with a value of $1$ if the corresponding features are active. Otherwise, they are initialized with a value of $0$.

The $Q-$values mentioned in equations \ref{TD_update} and \ref{Qlambda-TDerror} are stored in the form of weight vectors as:
\begin{equation}\label{Qvalue}
Q(s,a)=\sum_{i\in \mathcal F_{act}(s,a)}w_{i}
\end{equation}
where $F_{act}(s,a)$ is the set of active features for an agent in state $s$, taking an action $a$. A more detailed summary of the algorithm can be found in \cite{sutton1998reinforcement}.

Although off-policy methods such as the ones described above have been well known and widely used over the years, their use for autonomously handling multiple independent objectives has been limited, primarily owing to very few precedents on unsupervised identification of objectives in an agent's environment. Off-policy approaches with function approximation have also been known to have long standing issues with stability until recently \cite{sutton2011horde}. Although approaches for handling multiple independent objectives in parallel are rather limited, a number of multi-objective reinforcement learning approaches that handle multiple conflicting objectives exist. A comprehensive survey of such methods can be found in \cite{roijers2013survey}. 

The horde architecture of Sutton et al.\cite{sutton2011horde} has been shown to be able to learn multiple pre-defined objectives in parallel using independent reinforcement learning agents in an off-policy manner. The knowledge of these tasks is stored in the form of generalized value functions which makes it possible to obtain predictive knowledge relating to different goals of the agent. 
Modayil et al.\cite{modayil2014multi} and White et al.\cite{white2012scaling} also focus on learning multiple objectives in parallel using off-policy learning.
Apart from this, Sutton et al. \cite{Sutton98intra-optionlearning} used off-policy methods to simultaneously learn multiple options \cite{sutton1999between}, including ones not executed by the agent. They mention that the motivation for using off-policy methods is to make maximum use of whatever experience occurs and to learn as much as possible from them, which is an idea that is reflected in this work.  

In the works mentioned above, the multiple objectives that are learned in parallel are pre-defined. However, in this work, we focus on the case where the agent has no foreknowledge of the objectives in its environment. The objectives are identified by the agent itself via clustering.  Hence, the agent learns independently in the sense that as it moves through its environment, it identifies potential objectives and at least partially learns their associated value functions in parallel.

A similar approach is seen in Mannor et al. \cite{mannor2004dynamic}, where clustering is performed on the state-space to identify interesting regions. However, their approach was not online and the purpose of their work was to use these regions to automatically generate temporal abstractions.

We use a variant of the K-means clustering algorithm \cite{hartigan1979algorithm,anderberg2014cluster} to cluster features that are characteristic of secondary objectives. The approach is similar to that of Bhatia \cite{bhatia2004adaptive}, where an adaptive clustering approach is described. The difference lies in the fact that in our method, in addition to the mean, statistical properties such as the variance and number of members in each cluster are updated online and used for clustering as and when the environment is sensed by the agent. 

In general, the algorithm also bears similarities to some aspects of adaptive resonance theory  \cite{carpenter2016adaptive}. The procedure for finding and updating the winning cluster in our approach is similar to that for comparing input vectors to the recognition field, and updating recognition neurons towards the input vector in adaptive resonance theory. Perhaps the main differences in our approach are the nature and function of the threshold/vigilance parameter. In our approach, the threshold is related to the variance of the cluster, which varies dynamically as more members are acquired by the clusters. However, in both approaches, the threshold has an effect on the resolution of the clusters. Overall, our clustering approach is simpler, and it is only focused on being able to identify clusters in an online manner, without much consideration to factors such as biological plausibility. The details of the algorithm are discussed further in Section \ref{methodology}.


\section{Description}
\label{description}

\begin{figure}[H]
\centerline{\includegraphics[scale=0.43]{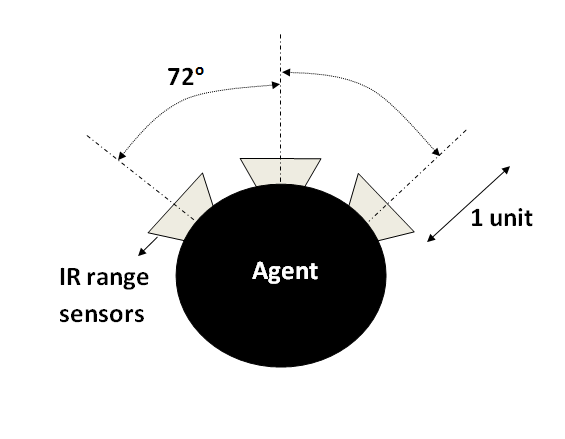}}
  \caption{The simulated agent and its range sensors}\label{fig:IR}
\end{figure}

\begin{figure}[H]
\centering
  \includegraphics[scale=0.3]{{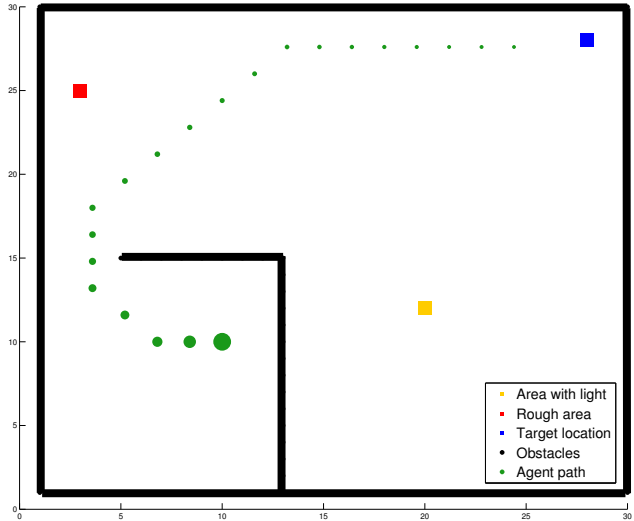}}
  \caption{Trajectory corresponding to one of the agent's policies to navigate to the target location in the simulated environment. The environment contains features such as a region with light, a rough region, obstacles and a target location}\label{fig:agentpath}
\end{figure}

In order to demonstrate the proposed approach for identifying and learning multiple objectives, we consider an agent in a 30x30 continuous space which contains obstacles, a region lit up by a light source, and a bumpy/rough area. We assume that characteristic features corresponding to these regions can be detected by the agent using its on-board sensors: a set of range sensors, a light detecting sensor, and an inertial motion unit (IMU) to sense changes in surface roughness. The range sensors on the robot are radially separated from each other by 72 degrees as shown in Figure \ref{fig:IR}, and are capable of sensing the presence of obstacles within 1 unit distance. A sample of the environment is shown in Figure \ref{fig:agentpath}.  

Initially, the agent has no foreknowledge of the environment, and can move forwards and backwards, sideways and diagonally up or down to either side. In addition to this, it can also hold its current position. Thus, a total of 9 deterministic actions (called the action set $\mathcal A$) are available for execution. These actions are executed sequentially according to the behavior policy, which depends on the primary objective assigned to the agent. The time step for action execution is set to be 200ms and the agent's velocity is set to be 8 units/s for the relevant actions. The features are a function of the environment and of the state of the agent, which is composed of the agent's \emph{(x,y)} position and its heading direction. Deriving these features from the agent's state is critical to learning, and is described below.

\subsection{Agent Features}
The agent is capable of sensing different features in the environment using its sensors. The sensors are simulated to have 5\% Gaussian white noise. We shall refer to the resulting feature vector as the \emph{environment feature vector} \emph{($\vec{F_{e}}$)}. For learning policies using linear function approximation, additional features for the agent's localization are needed. We shall refer to the vector of these features as the \emph{agent feature vector} \emph{($\vec{F_{a}}$)}. Hence, the full feature vector for the agent consists of both these feature vector components ($\vec{F}=\vec{F_{e}} \cup \vec{F_{a}}$).
All features used in this work are binary (1 or 0) for the sake of simplicity. 

 The feature vector \emph{$\vec{F_{e}}$} consists of the following:
\begin{enumerate}
  \item Feature indicating either the presence or absence of obstacles as seen by any of the three range sensors.
  \item Feature corresponding to the presence or absence of light
  \item Feature corresponding to rough or smooth floor surfaces, as reported by the IMU
  \item Feature indicating whether the agent lies within the range of the specified target location 
\end{enumerate}

The agent feature vector \emph{$\vec{F_{a}}$} is composed of $30$ binary features corresponding to each dimension in the 2-dimensional space. It is concerned with the localization of the agent, and is used for learning the required policies. In \emph{$\vec{F_{a}}$}, the feature value is equal to 1 for the agent's current position and 0 for all other positions in the space. Hence, the full feature vector consists of 64 (60 localization and 4 environment) feature elements. 

Only \emph{$\vec{F_{e}}$} is passed into the clustering algorithm to identify different regions of interest, whereas the full feature vector is used for the $Q-\lambda$ updates.

\section{Methodology}
\label{methodology}

Section \ref{description} described the simulated environment, the agent and the features it is capable of sensing. In this section, we describe the methodology used to identify regions of interest in the feature space and how these regions, treated as secondary objectives, can be learned using off-policy methods.

\subsection{Adaptive Clustering}
\label{cluster}
As described earlier, the feature vector sensed by the agent consists of features relating to the environment as well as features for localization of the agent. The agent is initially assigned an arbitrary primary objective, which is specified in terms of $\vec{\psi}$, which is a particular configuration of $\vec{F_{e}}$. In specifying $\vec{\psi}$, apart from the binary values that each feature can take, a `don't care' \tgknotes{don't care is more appropriate because it is a technical term used in digital logic} case is also included. During the task specification, if a primary objective feature is associated with the `don't care' case, it implies that any feature value sensed for that feature is considered acceptable during the search for $\vec{\psi}$ in the feature space. In learning the primary objective, the agent learns a policy that takes it from any arbitrary state in the environment to a state where $\vec{F_{e}}$ matches the feature vector described by $\vec{\psi}$

As the agent moves through the environment in search of the feature vector specified by $\vec{\psi}$, it is continuously presented with new $\vec{F_{e}}$ vectors. Our approach is to cluster these features as and when they are seen. The K-means \cite{romesburg2004cluster,anderberg2014cluster} algorithm is a simple and popular algorithm used for unsupervised clustering. However, it requires prior knowledge of the number of clusters present in the feature space. This does not suit our application, as we assume no prior knowledge about the environment. The algorithm was therefore modified in order to make it adaptive, so that new feature vectors that seem different from the others may `seed' new clusters. This is done by continuously updating the statistical properties such as the mean, variance and number of members in each cluster, and by measuring the closeness of new features $\vec{F_{e}}$ to the mean vectors of the different clusters.

Each new cluster that is seeded is initially set to have non-zero variance, which we shall refer to as the \emph{seed variance}. This is done in order to initially maintain a certain level of uncertainty about the cluster. The uncertainty reduces as more number of  samples are observed. As the agent moves through the environment, the Euclidean distance between the environment feature vector $\vec{F_{e}}$ that it sees, and the mean vector of each cluster is calculated, and the cluster corresponding to the minimum distance is chosen as the `winning' cluster. Next, the element-wise absolute distance between the mean of the winning cluster and $\vec{F_{e}}$ is computed. For each element, if this distance lies within `n' standard deviations of the mean of that feature element, then the $\vec{F_{e}}$ belongs to that cluster; if not, a new cluster is seeded. So `n' can be considered a tolerance parameter for the clustering algorithm. Each time a cluster receives a new member, the mean and variance of each of the $j^{th}$ feature element in the cluster is updated online using the corresponding elements of  $\vec{F_{e}}$. The straightforward equations governing the updates are defined by the first and second statistical moments of the sensor measurement in Equation \ref{meanupdate} and Equation \ref{varupdate} respectively.

\begin{equation}\label{meanupdate}
\mu_{j}\longleftarrow(N_{C}*{\mu}_{j}+F_{e}^{j})/(N_{C}+1)
\end{equation}

\begin{equation}\label{varupdate}
{\sigma_{j}}^2\longleftarrow (N_{C}*(\sigma_{j}^2 +\mu_{j}^2)+{F_{e}^{j}}^2)/(N_{C}+1)-\mu_{j}^2
\end{equation}

\begin{equation}\label{membersupdate}
N_{C}\longleftarrow N_{C}+1
\end{equation}
where $\mu_{j}$ and ${\sigma_{j}}^2$ are respectively the mean and variance of the $j^{th}$ feature element in the cluster, whereas $N_{C}$ is the number of members in cluster $C$. The structure of the clustering algorithm is summarized in Algorithm \ref{alg:algorithm1}.

 Overall, the algorithm serves to cluster the feature space in an unsupervised and adaptive manner without prior knowledge of the number of clusters that exist in the space. Each of the identified clusters is treated as a secondary objective which is learned in parallel with the primary objective using off-policy methods.

\begin{algorithm}[h!]
    \caption{Adaptive clustering algorithm}
  \begin{algorithmic}[1]
    \STATE \textbf{Inputs}: Feature vector $\vec{F_{e}}$, variance threshold parameter $n$, number of existing clusters $K$ (initially set to 1), existing clusters $C$ and their properties: mean $\vec{\mu}$, standard deviation $\vec{\sigma}$ (elements initialized with non-zero seed variance for a new cluster) and number of members $N_{C_{K}}$ (initialized to $1$ for a new cluster)
    \FOR {i=1:K}
    \STATE $d_{i}=Euclidean\_distance(\vec{F_{e}},\vec{\mu}_{i})$
    \ENDFOR
    \STATE $win=\{argmin(d)\}$
      \IF {$|(F_{e}^{j}-{{\mu}_{win}^{j}})| \geq n*{\sigma_{win}^{j}}$ for each feature $F_{e}^{j}$ in $\vec{F_{e}},$}
      \STATE $K=K+1$
      \STATE $\vec{F_{e}}\in C_{K}$
      \ELSE 
      \STATE {$\vec{F_{e}}\in C_{win}$}
      \STATE Update the mean and variance of each element in the winning cluster
      
      ${\mu_{win}^{j}}\longleftarrow(N_{C_{win}}*{\mu_{win}^{j}}+F_{e}^{j})/(N_{C_{win}}+1)$
      
      ${\sigma_{win}^{j}}^2\longleftarrow (N_{C_{win}}*({\sigma_{win}^{j}}^2 +{\mu_{win}^{j}}^2)+{F_{e}^{j}}^2)/(N_{C_{win}}+1)-{\mu_{win}^{j}}^2$
      \STATE Update the number of members in the winning cluster
      
      $N_{C_{win}}\longleftarrow N_{C_{win}}+1$
      \ENDIF
      
  \end{algorithmic}
  \label{alg:algorithm1}
\end{algorithm}

\subsection{Off-Policy Learning}
\label{Offpol}


The clustering algorithm described in Section \ref{cluster} groups feature vectors $\vec{F_{e}}$  into different clusters in an adaptive and unsupervised manner. As and when each new cluster is seeded, an associated set of weight vectors (to learn the corresponding $Q$ function) is also created. 
The mean vectors of each of these clusters is treated as a secondary objective feature vector  $\vec{\phi}$, and the associated set of weight vectors is updated using the $Q-\lambda$ algorithm, based on actions resulting from the behavior policy. So during each episode of learning of the primary objective, the secondary objectives identified by the clustering algorithm are learned simultaneously using off-policy learning. At the same time, new secondary objectives are identified by the clustering algorithm. If `M' secondary objectives are identified, the $Q-\lambda$ updates are performed `M' times in addition to the one time that the update is carried out for the primary objective. For these additional updates, the scalar rewards are dictated by the associated secondary objective. The reward structure used here is simple, and all objectives are treated equally. A reward of 100 is awarded for successfully achieving an objective, and a living penalty of 10 is associated with each step that does not correspond to the fulfillment of an objective. In addition to this, irrespective of the objective, a penalty of 100 is assigned for bumping into obstacles. More sophisticated reward structures that reflect the relative importance of the different objectives may be explored in the future.
The overall algorithm is summarized in Algorithm \ref{alg:algorithm2}.

\begin{algorithm}[H]
    \caption{Identifying and learning objectives using clustering and off-policy methods}
  \begin{algorithmic}[1]
    \STATE \textbf{Inputs}: Primary objective feature vector ($\vec{\psi}$), variance threshold parameter $n$, number of existing clusters K (initially set to $1$), starting state ($x_{start}$), weight vector  $w_{O}$, $Q-\lambda$ parameters for primary objective: discount factor ($\gamma$), learning rate ($\alpha$), exploration parameter ($\epsilon$), decay rate parameter for eligibility traces ($\lambda$) number of iterations for $Q-\lambda$ ($N\_iter$),  existing clusters $C$ and their properties: mean $\vec{\mu}$, standard deviation $\vec{\sigma}$ and number of members $N$
    \FOR {i=1:$N\_iter$}
    \STATE state=$x_{start}$
    \STATE $\vec{F_{e}}=getfeaturesfromstate(state)$
    \WHILE{$\vec{F_{e}} \neq \vec{\psi}$}
    \STATE Take $\epsilon$-greedy action and visit new state $x_{new}$
    \STATE $\vec{F_{e}}_{new}$=$getfeaturesfromstate$($x_{new}$) 
    \STATE Cluster $\vec{F_{e}}_{new}$ using algorithm \ref{alg:algorithm1}
    \IF {New clusters are formed,}
    \STATE Seed $w_{new\_cluster}$ and update K
    \ENDIF
     \IF {$\vec{F_{e}}_{new}==\vec{\psi}$,}
    \STATE	reward=high
    \ELSE {reward=low}
    \ENDIF
    \STATE Update $w_{O}$ using $Q-\lambda$ equations
    \FOR {j=1:K}
    \STATE $\vec{\phi}$=$\vec{\mu}_{j}$
    \IF {$\vec{F_{e}}_{new}==\vec{\phi}$}
    \STATE	reward(j)=high
    \ELSE reward(j)=low
    \ENDIF
    \STATE Update $w_{j}$ using $Q-\lambda$ equations
    \ENDFOR
    \STATE $x=x_{new}$
	\STATE $\vec{F_{e}}$=$\vec{F_{e}}_{new}$
    \ENDWHILE
    \ENDFOR
    
  \end{algorithmic}
  \label{alg:algorithm2}
\end{algorithm}

\section{Results}
\label{results}

\begin{figure}[H]
\centerline{\includegraphics[scale=0.31]{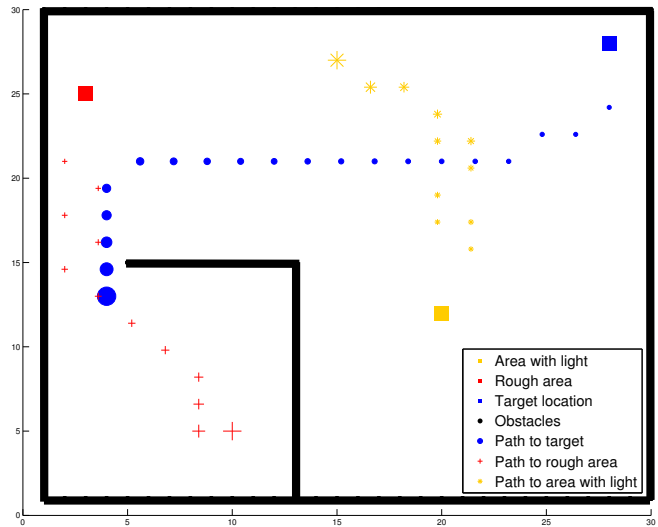}}
  \caption{Trajectories corresponding to the policies for different objectives learned by executing the behavior policy for the original task}\label{fig:allpols}
\end{figure}

In this section, we summarize the results obtained by applying the methodology described in Section \ref{methodology} to the agent and environment described in Section \ref{description}. The sample environment used for the simulations are shown in Figures \ref{fig:agentpath} and \ref{fig:allpols}. In these figures, larger markers corresponding to the agent's path signify points closer to the starting position of the agent. The configuration of the obstacles in the environment is set up to be similar to the `puddle world' problem \cite{sutton1996generalization}, in the sense that in order for the agent to navigate to the required location, it may need to temporarily move away from its target location. 

\begin{figure}[h!]
\centerline{\includegraphics[scale=0.7]{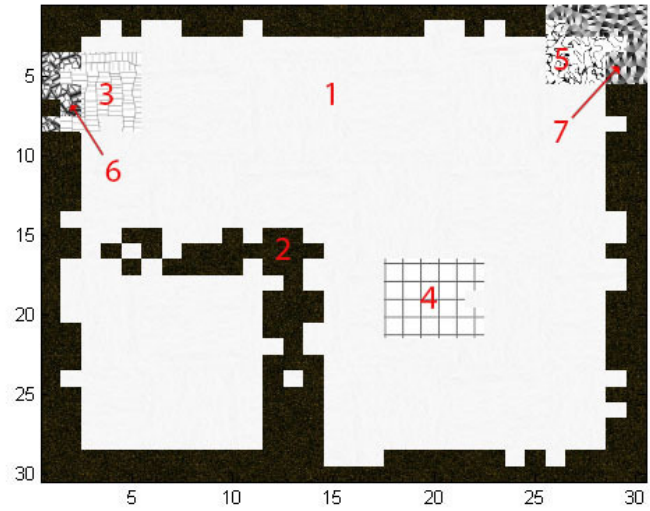}}
  \caption{Different clusters detected by the agent for the environment shown in Figure \ref{fig:agentpath}}\label{fig:clusters}
\end{figure}
\begin{figure}[H]
\centerline{\includegraphics[scale=0.6]{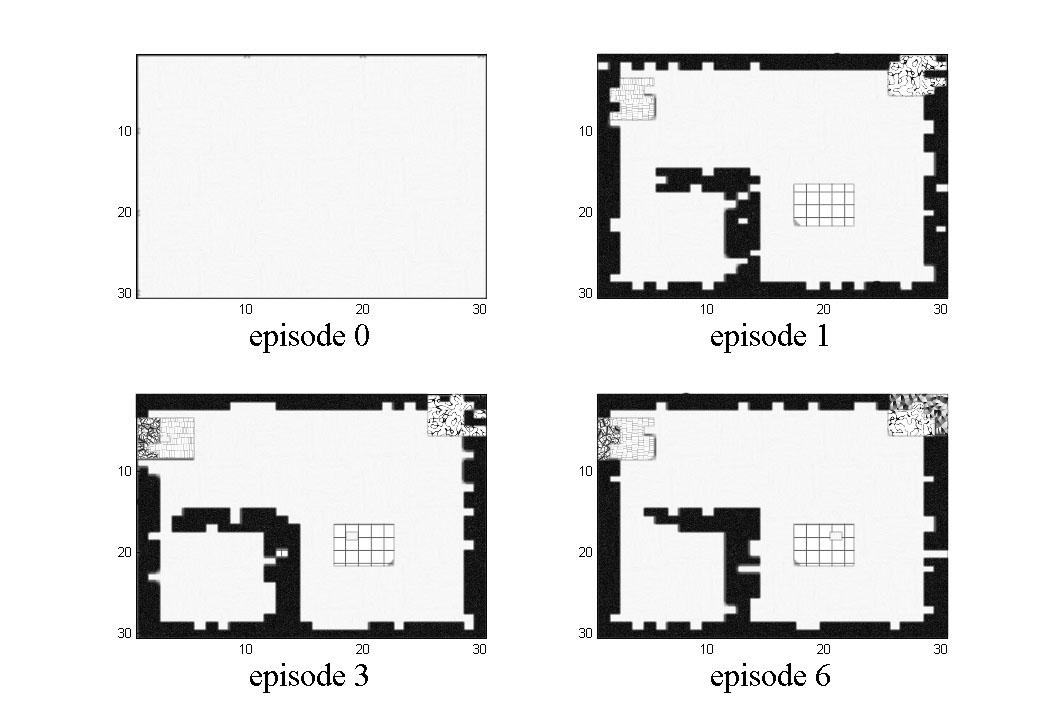}}
  \caption{Progression of cluster formation with episodes of the $Q-\lambda$ algorithm}\label{fig:clusterformation}
\end{figure}

The agent executes an $\epsilon$- greedy policy while learning a primary objective, during which it senses features $\vec{F_{e}}$ in its environment, and continuously sorts them into new or existing clusters as dictated by Algorithm \ref{alg:algorithm1}. Figure \ref{fig:clusters} shows the clusters identified by the algorithm after the $Q-\lambda$ algorithm is applied to learn the primary objective of navigating to the target location. In Figure \ref{fig:clusters}, a total of 7 clusters can be seen, each marked with a distinct texture and number. It is also seen that regions that have an overlap of different types of features are sorted as different clusters. For example, the region near the top right corner of Figure \ref{fig:clusters} contains a cluster (marked as cluster 7) which corresponds to the overlap between an area around the target location and the presence of an obstacle. 
In Figure \ref{fig:clusterformation}, it is seen that during episode 1, this overlapping area is not distinguished as a separate cluster. This changes as the episodes proceed, and the overlapping area is eventually identified as a distinct cluster after episode 6. A similar overlap exists (marked as cluster 6 in Figure \ref{fig:clusters}) around the area with high floor roughness near the top left corner of the environment. This shows that with a larger number of samples, the clustering algorithm is capable of distinguishing different combinations of feature elements in the feature space in an unsupervised manner.

\begin{table}[H]
\begin{scriptsize}
\begin{center}
\caption{Average number of clusters formed as clustering parameters seed variance and clustering tolerance (n) are varied}
\label{table:clusteringparams}
\scalebox{1.2}{
    \begin{tabular}{ | l | l | l | l | l | l | l }
    \hline
	&$n$=0.1  & $n$=1  &$n$=1.1  & $n$=1.5  & $n$=2 \\ \hline
	seed variance=0.1 & 6.82 & 6.65 & 1.93 & 1.36 & 1.39\\ \hline
    seed variance=1 & 6.77 & 6.33 & 1.47 & 1.19 & 1\\ \hline
     seed variance=100& 6.49  & 6.51 & 1.63 & 1.06 & 1\\ \hline
    \end{tabular}    
}
\end{center}
\end{scriptsize}
\end{table}

Table \ref{table:clusteringparams} shows the average number of clusters identified as the seed variance and the clustering tolerance $n$ are varied. The values shown are compiled for 50 $Q-\lambda$ runs with an exploration parameter $\epsilon=0.3$ for 1000 episodes. The other parameters are the learning rate $\alpha=0.3$, the discount factor $\gamma=0.9$ and the trace decay rate parameter $\lambda=0.9$. These parameters were kept constant for the $Q-\lambda$ runs.
The results shown in Table \ref{table:clusteringparams} suggest that the clustering is sensitive to the clustering tolerance, as we may have expected. The lower the value of $n$, the larger is the number of clusters identified. As per algorithm \ref{alg:algorithm1}, the condition for new clusters to be formed is:
\begin{equation}\label{noclusters1}
|(F-\mu)|\geq n\sigma
\end{equation}
where F is the value of the feature element and $\mu$ and $\sigma$ are the mean and standard deviation of the associated `winning' cluster. From Chebyshev's inequality, the probability of clusters forming is bounded by:
\begin{equation}\label{chebyshev}
P(|(F-\mu)|\geq n\sigma)\leq 1/n^2
\end{equation}
When $n\leq1$, the term on the right hand side of equation \ref{chebyshev} is $\geq1$. Since probabilities cannot exceed $1$, all cases of $n\leq1$ are equivalent in this sense.   When $n>1$, the probability reduces, and the clustering performance drops. This could provide some explanation for the trends seen in Table \ref{table:clusteringparams}.
It also suggests that the clustering tolerance $n$ should ideally be set to a value $\leq1$ if clusters are to be identified effectively.

In addition to this, the performance of the clustering algorithm is observed to be more or less independent of the seed variance. This is because the variance of each cluster is continuously updated with each visit to a state. As more samples are obtained, the initial seed variance assigned to a cluster is quickly corrected to be closer to its true value.
For the given environment and agent, the clusters were mostly identified during the early episodes of $Q-$ learning. Figure \ref{fig:clusterformation} shows a typical progression of cluster formation with the number of episodes. 

The clusters identified by the adaptive clustering algorithm are passed on as secondary objectives to be learned using off-policy learning. The mean vector of these clusters, which describe the features represented by the cluster are then used to construct the feature vectors of the respective secondary objectives ($\vec{\phi}$).  

For the case of feature vectors $\vec{F_{e}}$ with a large number of elements, the number of clusters identified is likely to be large. For example, when $60$ additional features were added to the environment feature vector described in Section \ref{description}, a total of $748$ different clusters were formed. In such cases, it may be more practical to choose a certain number of clusters based on some predetermined criteria, and learn their associated policies. An example of one such criterion would be the average value of the temporal difference (TD) error across the state-action space, with secondary objectives corresponding to lower average error values being preferred. The hypothesis is that since the reward structures for the different objectives are similar, objectives with the lowest average TD error are likely to have have been learned more reliably. Hence, the objectives could be sorted in this manner according to the reliability of their associated $Q-$functions.

Once the clustering algorithm identifies a secondary objective, its corresponding weight vectors are initialized, and its value function is learned by making use of whatever experience could be gained from the agent's behavior policy. At the end of each episode of learning, the agent's starting position is reset to a random non-goal state. Ideally, after $Q-$learning, the agent should be able to generate optimal trajectories starting from anywhere in its environment subject to the assumption that each state-action pair is visited infinitely often. Table \ref{table:epsilonvar} shows the percentage of starting positions, picked uniformly from the environment shown, which lead to acceptable policies. The variation of this quantity with the number of episodes and different exploration parameters is also shown. The policies being evaluated are generated by having the agent take a series of greedy actions (as per the value functions it has learned) till the goal state is reached. A policy (and its corresponding trajectory) is considered acceptable if the path resulting from it is similar to the one computed using the A-star algorithm \cite{4082128}, which computes the optimal trajectory between two points, given perfect information regarding the environment. Policies whose resulting paths are more than 50\% longer than those computed using A-star are not considered acceptable. The tolerance (50\%) used here may seem very high, but this is because the aim of our approach is to provide the value functions of the secondary objectives with good initializations using whatever experience occurs; it is not to learn the secondary objectives perfectly. We posit that given the same behavior policy, the agent will be able to learn the optimal value functions corresponding to the secondary objectives much faster when starting with good initial estimates. The comparison with A-star is performed in order to evaluate the general quality of the value functions for different objectives. Each of the clusters shown in Figure \ref{fig:clusters} is treated as a secondary objective, but only the values for meaningful secondary objectives such as navigating to the regions with light or those corresponding to a rough area have been tabulated.

\begin{table}[h!]
\begin{scriptsize}
\begin{center}
\caption{Percentage of starting positions picked uniformly from the environment, that lead to acceptable policies for the primary and selected secondary objectives }
\label{table:epsilonvar}
\scalebox{1.2}{
    \begin{tabular}{ | l | l | l | l | l | l |}
    \hline
	No. of episodes& Objectives & $\epsilon$=0.1 & $\epsilon$=0.3 &  $\epsilon$=0.7 & $\epsilon$=1 \\ \hline
	    100 & Primary objective  & 51.97 & 60.32 & 71.35 & 81.62 \\ 
    & Light area  & 12.59 & 17.84 & 26.70 & 72.37 \\ 
     & Rough area  & 15.43 & 16.78 & 28.52 & 60.48 \\ \hline
     300 & Primary objective  & 69.92 & 72.57 & 80.76 & 82.13 \\ 
         & Light area  & 12.65 & 18.38 & 29.03 & 80.05 \\ 
     & Rough area  & 16.16 & 18.27 & 27.41 & 81.94 \\ \hline
              1000 & Primary objective  & 78.62 & 82.35 & 88.24 & 90 \\
         & Light area  & 16.22 & 21.30 & 52.16 & 85.54 \\ 
     & Rough area  & 17.19 & 21.38 &  41.08 & 85.41 \\ \hline
    \end{tabular}    
}
\end{center}
\end{scriptsize}
\end{table}

The values from Table \ref{table:epsilonvar} suggest that in general, the value functions resulting from policies that are more exploratory in nature result in acceptable policies from a greater percentage of starting positions. This is expected, as more state-action pairs have a chance to be visited when $\epsilon$ is set to be large. In general, the  percentage also goes up marginally with an increased number of learning episodes. This is natural, as a larger number of episodes allow greater opportunity for more state-action pairs to be visited more frequently.

Figure \ref{fig:allpols} shows some of the sample learned trajectories for both the primary objective (navigating to the target location) as well as the two selected secondary objectives. The trajectories leading to the `light' and `rough' areas in Figure \ref{fig:allpols} correspond to policies that were learned by first identifying the relevant regions in the feature space as secondary objectives, and then simultaneously  learning (partially) their associated action-value functions through off-policy learning. 
If each of the `N' secondary objectives were to be learned sequentially using $Q-$learning, `N' additional phases of exploration and learning would have been required. Here, the value function for all the secondary objectives are learned at least partially from the experience gained while learning to perform the primary objective. 
Although the percentages in Table \ref{table:epsilonvar} seem to attain high values for the secondary objectives only under more exploratory behavior policies ($\epsilon=0.7$ and $\epsilon=1$), some knowledge of the corresponding objectives is gained even when the behavior policy is set to be relatively greedy. Even this partial knowledge of the secondary objectives could help provide some initial estimates of the value function when optimal policies corresponding to these objectives are required to be learned. In this manner, the efficiency of exploration is improved to some extent, irrespective of whether the agent's behavior policy is greedy or highly exploratory. 

\begin{figure}[h!]
\centerline{\includegraphics[scale=0.5]{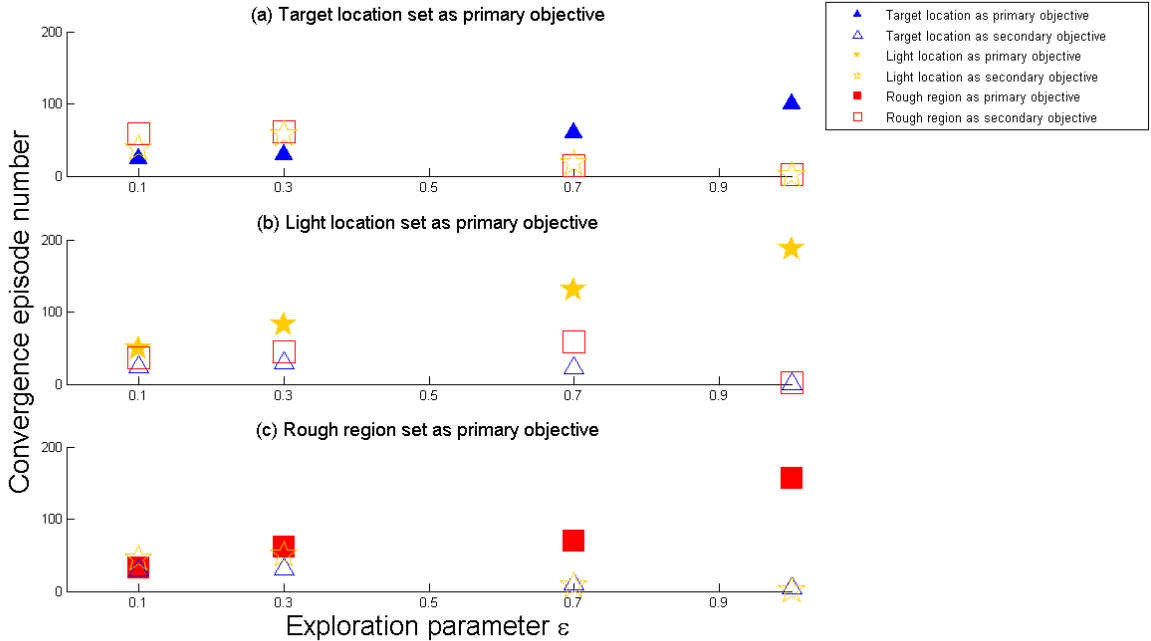}}
  \caption{Number of episodes for convergence for different values of $\epsilon$ with different objectives set as primary}\label{fig:results}
\end{figure}

This point is further emphasized through Figure \ref{fig:results}, where the number of episodes to convergence is measured and plotted for different objectives under different behavior policies.
Here, convergence is defined to be achieved when the agent is able to successfully navigate (have a trajectory length close to that specified by the A-star algorithm) to the required regions from the majority of locations in the environment.
In the simulations summarized by Figure \ref{fig:results}, different objectives are set as primary ones, and the agent is made to learn them while discovering and learning secondary objectives in parallel. Once the primary objectives converge, the partially converged weights of the secondary objectives are made to dictate the behavior policy. Then, the number of episodes for these secondary objective policies to converge is measured by checking for convergence after each episode. 
Each data point in Figure \ref{fig:results} is obtained by averaging the values of 50 $Q-\lambda$ runs with the corresponding $\epsilon$ values. The data points represented by solid shapes indicate primary objectives, whereas non-solid shapes represent secondary objectives.

From Figure \ref{fig:results} the convergence for secondary objectives is observed to be faster in comparison to the case where the same objective is learned from scratch as a primary one. This is because for the secondary objectives, learning is initialized with weight vectors that are already partially converged owing to the off-policy learning that occurred while learning the primary objective. 
For example, in Figure \ref{fig:results} (a) ($\epsilon$=0.7), the primary objective is set to be to navigate towards the target location. The corresponding value function weights converge in about 60 episodes. During this period, secondary objectives are simultaneously identified and learned. The partially converged weights for two of these objectives (light and rough region) are used to initialize the learning for the corresponding objectives till convergence. It is seen from Figure \ref{fig:results}(a) ($\epsilon$=0.7) that for the secondary objective of navigating to the rough region, convergence is achieved in only about 17 episodes on average. This is much faster than the case where the `rough region' objective is learned from scratch as a primary objective, where convergence takes place after about 70 episodes as indicted in Figure \ref{fig:results}(c) ($\epsilon$=0.7). For the secondary objective of navigating towards the light, convergence is achieved in 14 episodes (Figure \ref{fig:results}(a),($\epsilon$=0.7)), whereas it would have taken about 130 episodes (as seen from Figure \ref{fig:results}(b) ($\epsilon$=0.7)) if the light objective were to be learned from scratch as a primary objective. 

In general, for more exploitative behavior policies ($\epsilon$=0.1 and $\epsilon$=0.3 in Figure \ref{fig:results}), improvements may still exist, but it is less drastic. For example, when $\epsilon=0.3$, the objective of navigating to a region with light takes about 25 episodes fewer to converge when deployed as a secondary objective as compared to when deployed as a primary one. The reduced improvement is due to the fact that agents under a greedy policy seldom deviate much from their path towards the primary objective. As a result, secondary objectives are visited less frequently unless they happen to lie along the optimal path towards the primary objective. 


Hence, the effectiveness of the proposed methodology depends on the agent's behavior policy (whether exploitative or exploratory) as well as on the configuration of different objectives in the environment. However, in an arbitrary environment, our approach could possibly enable a significant reduction in the learning time (represented here by the number of episodes for convergence) required to learn the optimal policy for a secondary objective, even when relatively greedy policies are employed.

\section{Discussion}
\label{discussion}

As demonstrated in Section \ref{results}, the value function weights, even if partially converged, can make good starting points for carrying out subsequent $Q-$ learning episodes if improvement in the value function estimates is needed. 
Although we used the $Q-\lambda$ algorithm in this work, other off-policy methods could also be used, perhaps in conjunction with suitable abstraction techniques such as tile coding \cite{sutton1996generalization,whiteson2007adaptive}. 

In employing the approach described here, it is to be noted that the secondary objectives identified during clustering may or may not be of relevance to the agent in the future. Assessing the relevance or relative importance of these objectives could be an area for further research. In addition to this, the construction of the reward structure in a more informed manner could also be explored further.

Nevertheless, we believe our approach could be useful in several fields, with direct applications in transfer learning \cite{taylor_transfer_2009}, where it could provide jumpstart improvements \cite{lazaric2012transfer} when the partially learned weights are transferred within or across agents. It can also be useful in multi-agent applications \cite{tan1993multi,busoniu2008comprehensive}, as the value function information of the secondary objectives could be communicated to another agent whose primary objective is similar in nature to one of the original agent's secondary objectives. This could be a much more efficient approach, as each agent need not explore the environment from scratch. The exploration performed by other agents could be leveraged by subsequent agents to carry out their individual tasks.

\section{Conclusion}

The methodology developed and presented here demonstrates how the discovery and learning of potential objectives in an agent's environment is possible. Potential objectives are identified using an online, unsupervised and adaptive clustering algorithm. The identified objectives are then learned in parallel using off-policy methods. Both clustering as well as off-policy learning are demonstrated using a simulated agent and environment. The performance of the clustering algorithm with respect to its input parameters is tabulated and the findings are explained. The clustering algorithm is shown to be capable of identifying most of the distinct regions in the environment during the early episodes of  $Q-$ learning. Simulations conducted to validate the utility of this approach reveal that the agent is able to at least partially learn multiple objectives in parallel without any additional exploration. This is especially true when the behavior policy itself is exploratory in nature. The future scope, possible extensions to this work and its applications to fields such as transfer learning and multi-agent systems are also briefly discussed. 
Although the efficiency of our approach depends to some extent on the nature of the behavior policy and the configurations of objectives in the environment, we believe it presents a potential to dramatically improve the efficiency of exploration for reinforcement learning agents in unknown environments.

\section*{References}


\bibliography{RL_General,Abstraction,clustering,Transfer_Learning}


\end{document}